\documentclass[conference]{IEEEtran}
\IEEEoverridecommandlockouts

\usepackage{cite}
\usepackage{amsmath,amssymb,amsthm}
\usepackage{bbm}
\usepackage{algpseudocode}
\usepackage{graphicx}
\usepackage{textcomp}
\usepackage{mathtools}
\usepackage{xcolor}
\usepackage{algorithm}
\usepackage{bm}
\usepackage{subcaption}
\usepackage{array}
\usepackage{booktabs}

\def\BibTeX{{\rm B\kern-.05em{\sc i\kern-.025em b}\kern-.08em
    T\kern-.1667em\lower.7ex\hbox{E}\kern-.125emX}}
\usepackage{url}

\usepackage{tikz}
    \usetikzlibrary{shapes.arrows}
    
\usepackage{pgfplots}
\pgfplotsset{
compat=1.3,
legend style={font=\footnotesize, fill opacity=0.7,  draw opacity=1, text opacity=1, draw=white!15!black, legend cell align=left, align=left}, 
width=6cm, 
height=6cm,
yminorticks=false,
xminorticks=false,
title style={font=\small},
tick style={color=black},
tick label style={font=\small},
grid style={line width=.1pt, draw=gray!20},
major grid style={line width=.1pt,draw=gray!20},
}
\usepgfplotslibrary{colorbrewer}
\usepgfplotslibrary{groupplots}
\usepgfplotslibrary{fillbetween}
\usetikzlibrary{decorations.pathreplacing}
\usetikzlibrary{plotmarks}
\usepackage[]{fancyhdr}
\usetikzlibrary{patterns}
\pgfplotsset{every tick label/.append style={font=\footnotesize}}

\usepackage{tikz}
\usetikzlibrary{shapes, arrows, positioning}

\usepackage{subfig}
    
\usepackage{glossaries}
\newacronym{ml}{ML}{machine learning}
\newacronym{dl}{DL}{Deep Learning}
\newacronym{drl}{DRL}{Deep Reinforcement Learning}
\newacronym{rl}{RL}{reinforcement learning}
\newacronym{dqn}{DQN}{Deep Q-Network}
\newacronym{ddqn}{D-DQN}{Double Deep Q-Network}
\newacronym{dnn}{DNN}{deep neural network}
\newacronym{mdp}{MDP}{Markov Decision Process}
\newacronym{cnn}{CNN}{Convolutional Neural Network}
\newacronym{nn}{NN}{Neural Network}
\newacronym{ats}{ATS}{Adaptive Training Strategy}
\newacronym{pts}{PTS}{Periodic Training Strategy}
\newacronym{sjf}{SJF}{Shortest Job First}
\newacronym{mec}{MEC}{Mobile Edge Computing}
\newacronym{bs}{BS}{Base Station}
\newacronym{per}{PER}{Prioritized Experience Replay}
\newacronym{td}{TD}{Time Difference}
\newacronym{ai}{AI}{Artificial Intelligence}
\newacronym{cl}{CL}{Continual Learning}
\newacronym{qos}{QoS}{Quality of Service}

\newcommand{\mc}[1]{\mathcal{#1}}   
\DeclareMathOperator*{\argmax}{arg\,max}    

\algtext*{EndWhile}
\algtext*{EndIf}
\algtext*{EndFor}

\def \fwidth{\columnwidth}
\def \fheight {0.55\columnwidth}
\def \ffheight {0.5\columnwidth}

\definecolor{color0}{HTML}{0000ff}
\definecolor{color1}{HTML}{982bd6}
\definecolor{color2}{HTML}{c956af}
\definecolor{color3}{HTML}{e78088}
\definecolor{color4}{HTML}{f8ab5d}
\definecolor{color5}{HTML}{ffd700}
\definecolor{darkblue}{HTML}{00429D}
\definecolor{darkgreen}{HTML}{005c00}
\definecolor{gold}{HTML}{D4AF37}
\definecolor{darkred}{HTML}{910000}
\definecolor{darkslategray38}{RGB}{38,38,38}

\makeatletter

\title{To Train or Not to Train: Balancing Efficiency and Training Cost in Deep Reinforcement Learning for Mobile Edge Computing}

\author{\IEEEauthorblockN{Maddalena Boscaro, Federico Mason, Federico Chiariotti, Andrea Zanella}
\IEEEauthorblockA{Department of Information Engineering, University of Padova, Via G. Gradenigo 6/B, 35131, Padua, Italy \\
Emails: \{boscaromad, masonfed, chiariot\}@dei.unipd.it, andrea.zanella@unipd.it
\vspace{-0.5cm}}
\thanks{This work was partially supported by the European Union under the Italian National Recovery and Resilience Plan (NRRP) partnership on ``Telecommunications of the Future'' (PE00000001 - program ``RESTART''). F. Chiariotti's activities are funded by NRRP ``Young Researchers'' grant REDIAL (SoE0000009).}
}

\begin{document}

\maketitle

\begin{abstract}
\gls{ai} is a key component of 6G networks, as it enables communication and computing services to adapt to end users' requirements and demand patterns.
The management of \gls{mec} is a meaningful example of \gls{ai} application: computational resources available at the network edge need to be carefully allocated to users, whose jobs may have different priorities and latency requirements.
The research community has developed several \gls{ai} algorithms to perform this resource allocation, but it has neglected a key aspect: learning is itself a computationally demanding task, and considering free training results in idealized conditions and performance in simulations. In this work, we consider a more realistic case in which the cost of learning is specifically accounted for, presenting a new algorithm to dynamically select when to train the \gls{drl} agent that allocates resources.
Our method is highly general, as it can be directly applied to any scenario involving a training overhead, and it can approach the same performance as an ideal learning agent even under realistic training conditions.
\end{abstract}

\begin{IEEEkeywords}
Mobile Edge Computing, Deep Reinforcement Learning, Cost of Learning, Continual Learning
\end{IEEEkeywords}

\glsresetall

\section{Introduction}
\label{sec:intro}

\begin{tikzpicture}[remember picture, overlay]
      \node[draw,
      minimum width=4in] at ([yshift=-1cm]current page.north)  {This paper has been submitted to IEEE International Conference on Communications 2025.};
\end{tikzpicture}

The 6G paradigm will revolutionize mobile networks by integrating communication and computing in a holistic fashion, offering specialized services that constantly adapt to the specific needs of end users~\cite{zuo2023survey}.
This extreme customization of network applications will be allowed by the wide diffusion of \gls{ai}, which will become a core component of network management~\cite{jiao2024}.
The \gls{ai} native nature of 6G will therefore transform networks into proactive entities that respond to user requests while pursuing various optimization goals, far beyond human capabilities. 

A crucial use case of \gls{ai} in 6G is the orchestration of computational jobs in \gls{mec}~\cite{feng2022computation}. 
The offloading of tasks to \gls{mec}-capable \glspl{bs} will support autonomous vehicles, holographic communication, and many other breakthrough services, which require not only a huge bandwidth for data transmission, but also computationally intensive operations with strict timing deadlines~\cite{quy2023innovative}.
The high system load that these tasks will impose makes \gls{ai} algorithms necessary for automatically handling the allocation of computational resources, prioritizing critical applications and adapting to network dynamics without human intervention~\cite{ma2023dynamic}. 

Over the past decade, the benefits of \gls{ai} for \gls{mec} optimization have been highlighted in many circumstances~\cite{wang2020machine, zhou2021deep}.
In particular, the \gls{drl} paradigm can find decision-making strategies to allocate computational resources to the pending jobs, outperforming traditional heuristic approaches like \gls{sjf}~\cite{ru2013empirical}, which do not take into account the specific demand patterns of each \gls{mec} server. 
Traditional \gls{drl} frameworks assume that the target scenario is stationary in time, which enables the use of a pre-existing dataset for training and does not require modifying the learning model after its deployment.
This assumption is ill-suited to the expected requirements for 6G networks, which make extreme flexibility one of the key points of their operation~\cite{bagaa2022toward}. 
This then requires adopting a \gls{cl} approach, so that the \gls{drl} agents constantly acquire new information and retrain when the environment is subject to significant changes~\cite{wang2024comprehensive}. 

However, updating learning agents with new information makes it necessary to devote resources to train the agents themselves.
As learning is a computationally intensive operation, the optimization of \gls{cl} systems can have a strong impact on the performance of networks and computing facilities. 
In traditional \gls{drl} problems, the agent is considered as being outside the environment: its training and decision-making are assumed to be instantaneous, with no impact on the environment evolution.
However, this assumption becomes unrealistic in \gls{cl} systems, in which \gls{drl} training is both computationally expensive, requiring the computation of gradients over batches of experience samples, and online.

In this manuscript, we consider the problem of allocating computational jobs in a \gls{mec} server via a \gls{drl} agent.
Following our previous work~\cite{lahmer2024}, we define the \emph{cost of learning} as the overhead incurred by the allocation of \emph{training jobs} to improve the agent policy.
Since the ultimate goal of the agent is to maximize the \gls{mec} efficiency and serve as many users as possible within their deadlines, this poses a dilemma: should we use resources to train, and improve the agent accuracy in the long term, or maximize the number of resources assigned to the users, i.e., the immediate reward?
This trade-off represents a unique challenge that is often neglected in the \gls{drl} literature.

One potential solution is to implement a new learning agent with the goal of allocating \gls{mec} resources between regular and training jobs.
However, this would require additional resources to train the new agent, shifting the problem to the decision on when to train this meta-agent.
Another possibility is to include the decision directly in the action space of the agent.
In this scenario, the agent would need to learn not only how to allocate resources for generic incoming jobs but also for its own training jobs, identifying states in which learning is more beneficial.
However, as the policy improves over time, the priority of the learning decreases and the reward associated with the policy changes.
The result leads to a non-stationary environment, which potentially prevents the agent's policy from converging to the optimal solution.

In this work, we manage user and training resources separately, defining two heuristic strategies to decide when to allocate training jobs. 
The first is a periodic strategy that allocates training jobs according to a regular time frame.
The latter is an adaptive strategy that aims to estimate the most convenient states to allocate \gls{mec} resources for the training.
Unlike our previous work~\cite{lahmer2024}, which relied on strategies that needed to be designed \emph{ad hoc} for a specific application, these strategies are entirely general and are not limited to specific features of the \gls{mec} problem, but rather apply to any \gls{drl} setting in which cost of learning is an active concern.

Our major contributions are the following:
\begin{itemize}
    \item we implement a \gls{drl} agent to schedule job requests in a \gls{mec} environment with two classes of user requests with different priorities;
    \item we compare the \gls{drl} agent against a \gls{sjf} benchmark in an online training settings, highlight the impact of the cost of learning over the system performance;
    \item we design a novel heuristic algorithm, named \gls{ats}, to dynamically optimize the training process according to the estimated cost of learning in each possible state of the \gls{mec} server;
    \item we compare the proposed heuristic against a benchmark that does not consider the \gls{mec} state to decide when to take new training actions.
\end{itemize}
Aside from obtaining a significant performance advantage with respect to both naive \gls{cl} strategies and non-data-driven solutions, the proposed \gls{ats} algorithm is also fully general: unlike our previous work~\cite{lahmer2024}, which was an \emph{ad hoc} heuristic that considered specific features of the scenario, \gls{ats} uses the reward estimates made by the \gls{drl} agent itself to decide when and whether to train. This means that it can be applied directly to other cost of learning problems, including \gls{mec} systems with different statistics.

\section{System Model}
\label{sec:model}
We consider a \gls{mec} server connected to a cellular \gls{bs} and equipped with various types of computational resources that we consider as a unified pool with capacity $C$.
We assume that the system operates with discrete time slots $\tau$ and that a \gls{mec} scheduler allocates the available resources at the beginning of each slot.
We also assume that the \gls{bs} area covers $N_{\text{user}}$ users, each of which generates \emph{computing jobs} according to a Bernoulli process $\mathcal{X}_n \sim \mc{B}(p)$.
We define $p = \rho / (\mu N_{\text{user}})$, where $\rho$, named \emph{average load}, is the average fraction of the cluster capacity $C$ requested by the users, and $\mu$ is the average number of resources each job requires.

Each job $j$ requires a certain amount of resources $c_j$ and a fixed execution time $e_j$, which must be allocated when it is scheduled. It also has a fixed deadline $T_{j}$, i.e., the maximum time that the job can wait before execution without violating the user \gls{qos} requirements.
Thus, if a job request is not satisfied before the deadline expires, it is discarded by the \gls{mec} server. We assume that, at the beginning of each slot, the \gls{mec} scheduler can allocate resources to a single new job. Once \gls{mec} resources are assigned to a job, they remain allocated to it until its completion.

We can then formulate the job scheduling problem as an \gls{mdp}, following the framework proposed in \cite{staterepresentation}.
According to this model, the \gls{mec} server is provided with a finite buffer that can contain up to $L$ jobs. 
If a new job arrives when the buffer is already full, it is discarded by the system and marked as failed. 
The state of each job in the buffer is encoded by a tuple $\langle e_j,c_j,w_j,T_j\rangle$, where $w_j$ is the time that job $j$ has already spent in the buffer.
Hence, the buffer state can be represented by a $4 \times L$ matrix $\mathbf{B}$.

The full state of the \gls{mec} server depend on both the buffer conditions and the already allocated computational resources: we can represent the reserved resources for the next $M$ slots as a vector $\mathbf{g}\in\{0,1,\ldots,C\}^{M}$, where $g(m)$ indicates the amount of allocated resources at slot $m$. At the end of each time step, we set $g_{t+1}(m)=g_t(m+1)\,\forall m\in\{1,\ldots,M\}$, and $g_{t+1}(M)=0$. This notation is a more compact representation of the binary matrix defined in~\cite{staterepresentation}.
The time evolution of the buffer is rather simple: while $e_j$, $c_j$, and $T_j$ remain constant for each job in the buffer, $w_j$ is incremented by $1$ at each time slot, and the job is discarded if $w_j>T_j$.
If the \gls{mec} scheduler allocates a job on the server, it is removed from the buffer and marked as successfully executed.

At the beginning of each slot, the agent observes the system state $s=\langle\mathbf{B}, \mathbf{g}\rangle \in \mathcal{S}$ and chooses an action $a \in \mathcal{A}$. The action space is defined as $\mathcal{A} = \{0, 2, ..., B-1\} \cup \{ \varnothing \}$, where $B$ is the buffer capacity, $a = i$ indicates scheduling job in position $i$, while $a = \varnothing$, i.e., the \emph{void action}, indicates that no job is scheduled in the current time slot. Therefore, in each time unit, the learning agent can choose from $B+1$ possible actions.
We observe that some actions might be invalid, i.e., the chosen position in the buffer may be empty, or the job may require the allocation of resources that are already busy. Invalid actions are equivalent to the void action, while valid actions make jobs be immediately assigned to \gls{mec} resources.

We now introduce a delay penalty function $\phi(w,T)$, which describes the satisfaction level of a job with a deadline $T$ entering service after waiting $w$ slots in the buffer:
\begin{equation}
  \phi(w,T)=\begin{cases}
              1, &\text{if }w<\frac{T}{2};\\
              2\left(1-\frac{w}{T}\right), &\text{if }\frac{T}{2}\leq w<T;\\
              0, &\text{if }w\geq T.
            \end{cases}
\end{equation}
Hence, jobs are fully satisfied if they are served within $\frac{T}{2}$, after which the utility decreases linearly with delay until the deadline is reached.
Additionally, we introduce function $D(s,a)$, which counts the number of jobs in the buffer whose deadline expires if the scheduler chooses action $a$ in state $s$:
\begin{equation}
  D(s,a)=\sum_{j=1}^B I(w_j+1>T_j)I(c_j>0)I(j\neq a),
\end{equation}
where $I(\cdot)$ is the indicator function, whose value is $1$ if the condition in the argument is true and $0$ otherwise. 

We define the reward associated to state $s$ and action $a$ as
\begin{equation}
  r(s,a)=\begin{cases}
                    e_a\phi(w_a,T_a)-\sigma D(s,a), &\text{if }a\text{ is valid;}\\
                    -\sigma D(s,a), &\text{otherwise;}
                  \end{cases}
\end{equation}
where $\sigma \in \mathbb{R}^+$ is a tuning parameter. Scheduling a job right before its deadline does not provide a positive reward, as $\phi(T_j,T_j)=0$, but still prevents the job from being discarded, thus avoiding increasing the penalty term. The reward given to a job is also weighted by its duration, to consider the fact that longer jobs occupy the \gls{mec} server for longer.

In this work, we address the scheduling problem by a \gls{drl} approach, training a learning agent to maximize the long-term reward by associating each state $s \in \mathcal{S}$ with the optimal action $a^* \in \mathcal{A}$.
In particular, we consider the \gls{ddqn} algorithm~\cite{van2016deep}, which is an extension of traditional Q-learning, using two distinct neural networks, named \emph{policy} and \emph{target} networks.
While the policy network $Q(\cdot)$ is used to select actions, the target network $ \hat{Q}(\cdot)$ is used to estimate their associated Q values.
Our implementation also exploits \gls{per}, which allows the \gls{drl} algorithm to process more frequently the experience associated with higher reward during the training phase. We set the \gls{per} weight for each sample $i$ as
\begin{equation}
    \alpha_i = \exp \left(r_i-\max_{j \in \mathcal{E}}r_j\right),
\end{equation}
where $\mathcal{E}$ is the set containing the agent's past experience. 
Finally, we used $\varepsilon$-greedy exploration, changing the value of $\varepsilon$ according to a reverse sigmoid function scaled to match the number of episodes in the training.

\section{Cost of Learning Framework}
\label{sec:cost}

In real scenarios, \gls{mec} scheduling policies need to continuously adapt to time-varying demand patterns and user requirements.
In the case of learning-based optimization, as the one we consider, this requires to adopt a \gls{cl} approach in which the \gls{mec} scheduler is trained in real time.
Hence, we must take into account the computing overhead due to the \emph{cost of learning}, i.e., the need to allocate part of the \gls{mec} resources to the \emph{training jobs} used for the agent optimization.

In the following, we assume that the \gls{ddqn} algorithm can improve the \gls{drl} agent by processing $B$ batches of experience samples only when a training job is allocated by a meta-scheduler.
This latter is separated from the agent itself to avoid the convergence issues we discussed in the introduction.
We then consider a training job to require $c_{\text{tr}}$ resources for a single time slot, and present two heuristics that are able to dynamically decide when to allocate training jobs without harming the performance of the \gls{mec} system.

\subsection{\acrfull{pts}}
The first heuristic we propose is named \gls{pts} and schedules training jobs at regular intervals, referred to as \textit{training periods} and denoted as $T_{\ell}$.
At the start of each training period, a training job is immediately scheduled in the cluster, regardless of the system state.
Determining the optimal $T_{\ell}$ is then a key issue: lower values accelerate the training process, but also increase the load on the system, making the agent's task more difficult and taking computing resources away from the users.
On the other hand, if $T_{\ell}$ is too high, the learning algorithm converges too slowly, using a suboptimal policy for extended periods.

\subsection{\acrfull{ats}}

The major limit of \gls{pts} is that it operates independently of the current conditions of the \gls{mec} server.
To address this limitation, we design \acrfull{ats}, which aims at identifying when to allocate training resources.
The \gls{ats} algorithm leverages the current estimates of the Q-values to determine which states are the best for training, enabling the system to make more informed decisions. Indeed, states with higher Q-values are, by definition, more favorable to the agent: we can reasonably expect states with a higher expected long-term reward to be more resilient to the disruption caused by reserving resources to the training.

\begin{figure}[t]
\centering
\tikzset{
block/.style    = {draw, rectangle, minimum height = 2cm, minimum width = 1em}}
\begin{tikzpicture}[auto]

\node[rectangle, minimum height=0.25cm, minimum width=1.5cm, fill=white!50!blue] at (0.75,2.375) {};
\node[rectangle, minimum height=0.5cm, minimum width=1.25cm, fill=white!50!blue] at (0.625,2) {};
\node[rectangle, minimum height=0.25cm, minimum width=1.5cm, fill=white!50!blue] at (0.75,2.375) {};
\node[rectangle, minimum height=0.25cm, minimum width=0.75cm, fill=white!50!blue] at (0.375,1.625) {};
\node[rectangle, minimum height=0.25cm, minimum width=1.75cm, fill=white!50!blue] at (0.875,1.375) {};
\node[rectangle, minimum height=0.5cm, minimum width=1.5cm, fill=white!50!blue] at (0.75,1) {};

\foreach \x in {1,...,7}
    \draw[-,white!50!black] (0.25*\x,2.5) -- (0.25*\x,0);
\foreach \x in {1,...,9}
    \draw[-,white!50!black] (0,0.25*\x) -- (2,0.25*\x);

\node[rectangle,draw,minimum height=2.5cm, minimum width=2cm] at (1,1.25) {};

\node[rectangle, minimum height=0.25cm, minimum width=1.5cm, fill=white!50!blue] at (4.75,2.375) {};
\node[rectangle, minimum height=0.5cm, minimum width=1.25cm, fill=white!50!blue] at (4.625,2) {};
\node[rectangle, minimum height=0.25cm, minimum width=1.5cm, fill=white!50!blue] at (4.75,2.375) {};
\node[rectangle, minimum height=0.25cm, minimum width=0.75cm, fill=white!50!blue] at (4.375,1.625) {};
\node[rectangle, minimum height=0.25cm, minimum width=1.75cm, fill=white!50!blue] at (4.875,1.375) {};
\node[rectangle, minimum height=0.5cm, minimum width=1.5cm, fill=white!50!blue] at (4.75,1) {};
\node[rectangle, minimum height=0.25cm, minimum width=2cm, fill=white!50!red] at (5,0.875) {};

\foreach \x in {1,...,7}
    \draw[-,white!50!black] (4+0.25*\x,2.5) -- (4+0.25*\x,0);
\foreach \x in {1,...,9}
    \draw[-,white!50!black] (4,0.25*\x) -- (6,0.25*\x);

\node[rectangle,draw,minimum height=2.5cm, minimum width=2cm] at (5,1.25) {};

\draw[->] (2.25,1.25) to node[midway,above]{\small Training job}(3.75,1.25);

\node[anchor=south] at(1,2.5) {\small$s$};
\node[anchor=south] at(5,2.5) {\small$s^*$};

\node[rectangle, draw, minimum height=0.25cm, minimum width=0.25cm, fill=white!50!blue] at (6.4,1) {};
\node[rectangle, draw, minimum height=0.25cm, minimum width=0.25cm, fill=white!50!red] at (6.4,0.6) {};
\node[anchor=west] at (6.5,0.95) {\footnotesize User resources};
\node[anchor=west] at (6.5,0.55) {\footnotesize Training res.};
\node[rectangle, draw, minimum height=1cm, minimum width=2.3cm] at (7.3,0.8) {};

\end{tikzpicture}
\caption{Insertion of a training job in the resource grid.}
\label{fig:training}\vspace{-0.3cm}
\end{figure}
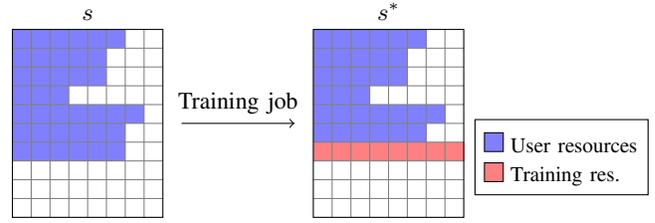

At the beginning of each slot, \gls{ats} simulates the effects of inserting a training job into the server, which would move the system from state $s$ to state $s^*$ (also called \textit{training state}), as shown in Fig.~\ref{fig:training}.
This can be simply done by marking $c_{\text{tr}}$ resources as occupied at the earliest possible moment.
Afterwards, \gls{ats} compares the Q-values for both the current and training state, getting
\begin{equation}
    \psi(s) = \max_{a  }{Q(s^*,a)} + \beta \left(\max_{a}{Q(s^*,a)} - \max_{a}{Q(s,a)}\right),
\end{equation}
where $\beta \in \mathbb{R}^+$ is a parameter that balances the contributions of the two equation terms.
The first term represents the maximum Q-value over all possible actions for the training state $s^*$, reflecting the potential reward of taking the optimal action in that state.
The second term measures the difference between the maximum Q-value of state $s^*$ and state $s$.
This difference intuitively quantifies the penalty (or cost) on the expected reward when a greedy action is performed in state $s^*$.
In essence, $\psi(s)$ measures the effect of inserting a training job on the expected long-term return, i.e., the cost of learning.

During the training process, \gls{ats} records the value of $\psi$ for each state in a dedicated buffer $\Psi$. Once $\Psi$ is full, we compare the new value of $\psi$ with the values in the buffer.
If $\psi$ is over the $99$th percentile, meaning the present state is more favorable for training than $99\%$ of recent states, a training job is allocated for the current time slot, and the network is trained using \acrshort{ddqn} algorithm.
Naturally, such a threshold should be tuned depending on the specific features of the problem as in different learning environments we may desire more or less intensive training phases.
On the other hand, no architectural changes to the \gls{ats} algorithm are needed.

During the early stages of the learning process, the Q-values may not be able to provide an accurate estimate of the state quality, which makes \gls{ats}'s decisions inaccurate.
To address this, we implement a method to determine when it is appropriate to use \gls{ats} to decide whenever to allocate training jobs.
Practically, during training, we record not only the value of $\psi$, but also the \gls{td} error $\delta$ and a newly defined measure $\phi$.
The values of $\delta$ and $\phi$ are used together to determine the moment in which the $\psi$ statistics become reliable.
In the case of \acrshort{ddqn}, the value of $\delta$ at time $t$ is
\begin{equation}
    \delta_t(s,a, s') = r(s,a) + \gamma\hat{Q}_t(s', \argmax_{a'}Q_t(s', a') ) - Q_t(s,a),
\end{equation}
where $\hat{Q}_t$ is the target Q-network at time $t$, and $s'$ is the agent observation at the subsequent slot.
When the value of the \gls{td} error becomes smaller than the difference $\phi(s)$ between the maximum Q-value for training state $s^*$ and the average Q-value across all possible actions for state $s$, the Q-values are sufficiently reliable to use \acrshort{ats}:
\begin{equation}
    \delta_t (s,a,s') \leq \phi(s)= \max_a{Q_t(s^*,a)} - \mathbb{E}_a[Q_t(s,a)] \text{.}
\end{equation}
In contrast, if the \gls{td} error exceeds $\phi(s)$, the Q-values may be less reliable or more uncertain. In such cases, it is advisable to go back to \acrshort{pts}. Indeed, the periodic alternative, not relying on Q-value estimates, can avoid the potential risks associated with inaccuracies in Q-value estimates.

\section{Simulation Settings and Results}
\label{sec:results}

In this section, we discuss the results of the simulations we performed to evaluate the proposed cost of learning framework in the MEC system described in Sec. \ref{sec:model}.
We compare the proposed \acrshort{pts} and \acrshort{ats} approaches in both stationary and dynamic environments.
We also consider two benchmarks, namely the classical \gls{sjf} algorithm~\cite{bruno1976sequencing}, which is commonly used as a standard approach for resource allocation, and an ideal \gls{drl} solution with no cost of learning, which provides an upper bound for \gls{drl} performance in realistic settings. 

In our simulations, we consider a \gls{mec} server with $C=20$ computing resources, serving users that have two kinds of jobs, that we call \emph{short} and \emph{long}. 
Short jobs have an execution time $e_{\text{short}} \sim U(\left\{\frac{C}{20},\ldots,\frac{3C}{20}\right\})$, while long jobs have an execution time $e_{\text{long}} \sim U(\left\{\frac{2}{5}C,\ldots,\frac{3}{5}C\right\})$.
The computational requirement $c$ is the same for both classes, $c \sim U(\left\{\frac{C}{4},\ldots,\frac{C}{2}\right\})$.
Jobs are generated independently at each user, with a probability $p_{\text{short}}=0.2$ of being short. The \gls{mec} load is then
\begin{equation}
\rho=\frac{3}{8}N_{\text{user}}p\left(\frac{1}{2}-\frac{2p_{\text{short}}}{5}\right),
\end{equation}
where $N$ and $p$ are the number of users and the Bernoulli probability associated to a single user, as defined in Sec.~\ref{sec:model}.

Finally, each training job takes up all the \gls{mec} resources when entering the server, i.e., $c_{\text{tr}} = C$. We use the Adam optimizer~\cite{zhang2018improved} to update the Q-networks' parameters, considering a maximum learning rate of $\alpha=10^-3$.
Table \ref{tab:1} summarizes the main parameters of our system, considering both the definition of the environment and the learning algorithm.

\begin{figure}[t]
\begin{table}[H]
\centering
\begin{tabular}{lcc}
\toprule
Parameter & Symbol & Value \\ \midrule
Average load & $\rho$ & $0.1-0.3$ \\
Size of the job buffer & $L$ & $10$ \\
Server computational capacity & $C$ & 20\\
Allocation horizon & $M$ & 20\\
Maximum waiting time & $T_{\text{short}}, T_{\text{long}}$ & $4, 8$ \\
User number & $N_{\text{user}}$ & $1000$ \\ \midrule
Number of training episodes & $N_{\text{train}}$ & $1000, 1500$ \\
Number of testing episodes & $N_{\text{test}}$ & $100$ \\
Number of slots per episode & $N_{\text{slot}}$ & $1000$ \\
Discount factor & $\gamma$ & $0.95$ \\
Batch size & $b$ & $16$ \\
Number of batches per training job & $B$ & $10$ \\
Weight soft update parameter  & $\tau$ & $0.005$ \\
Learning rate of Adam optimizer & $\alpha$ & $10^{-3}$ \\
ATS tuning parameter & $\beta$ & 0.4 \\ 
\bottomrule
\end{tabular}
\caption[Parameters of the system.]{Parameters of the system}
\label{tab:1}
\end{table}\vspace{-0.6cm}
\end{figure}

\subsection{Stationary scenario}
Firstly, we consider a stationary \gls{mec} scenario, where the average load is fixed and equal to $\rho=0.3$.
We simulate the resource allocation system over $N_{\text{train}}=1000$ episodes, each consisting of $N_{\text{slot}}=1000$ time slots. When using learning-based methods, we employ the $\varepsilon$-greedy strategy with exponential decay for the first $350$ episodes and then maintain a constant value $\varepsilon=0.1$ for the remaining episodes.

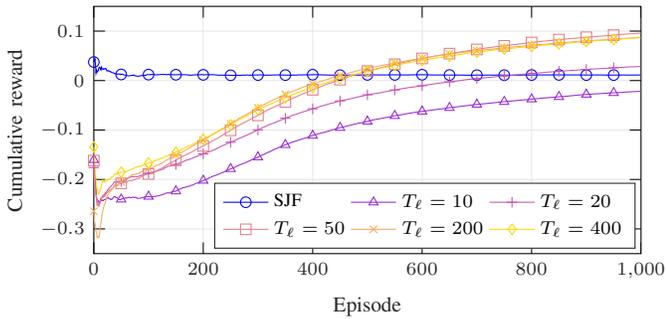
\begin{figure}[t]
\centering
\begin{tikzpicture}
\begin{axis}[%
width=\fwidth,
height=\fheight,
legend style={legend cell align=left, fill opacity=0.6, draw opacity=1, text opacity=1, legend columns=3, align=left, draw=white!15!black, font=\scriptsize, at={(0.99, 0.02)}, anchor=south east},
xlabel style={font=\footnotesize\color{white!15!black}},
ylabel style={font=\footnotesize\color{white!15!black}},
tick label style={font=\scriptsize\color{white!15!black}},
xmajorgrids,
ymajorgrids,
xmin=0,
xmax=1000,
ytick={-0.3,-0.2,-0.1,0,0.1},
xlabel={Episode},
ymin=-0.35,
ymax=0.15,
ylabel={Cumulative reward},
axis background/.style={fill=white}
]
\addplot [color=color0, mark=o, mark repeat=50, mark options={solid}]
table[x=Episode,y=SJF]{./fig/periodic_cumulative.csv};
\addlegendentry{SJF}

\addplot [color=color1, mark=triangle, mark repeat=50, mark options={solid}]
table[x=Episode,y=T10]{./fig/periodic_cumulative.csv};
\addlegendentry{$T_{\ell}=10$}

\addplot [color=color2, mark=+, mark repeat=50, mark options={solid}]
table[x=Episode,y=T20]{./fig/periodic_cumulative.csv};
\addlegendentry{$T_{\ell}=20$}

\addplot [color=color3, mark=square, mark repeat=50, mark options={solid}]
table[x=Episode,y=T50]{./fig/periodic_cumulative.csv};
\addlegendentry{$T_{\ell}=50$}

\addplot [color=color4, mark=x, mark repeat=50, mark options={solid}]
table[x=Episode,y=T200]{./fig/periodic_cumulative.csv};
\addlegendentry{$T_{\ell}=200$}

\addplot [color=color5, mark=diamond, mark repeat=50, mark options={solid}]
table[x=Episode,y=T400]{./fig/periodic_cumulative.csv};
\addlegendentry{$T_{\ell}=400$}

\end{axis}
\end{tikzpicture}%
\caption{Cumulative reward during the training as a function of the training period $T_{\ell}$ in the static scenario.\label{fig:cumulative_T}}\vspace{-0.3cm}
\end{figure}

We first study the effect of the length of the training period using \acrshort{pts}, experimenting with different values to identify the most effective configuration.
Fig.~\ref{fig:cumulative_T} shows the cumulative average reward, i.e., the cumulative sum of the rewards accumulated in the training phase.We note that longer training period values lead to a much better performance during the training, quickly overcoming the initial disadvantage with respect to \gls{sjf}. As training jobs require all the \gls{mec} server's resources for a single time slot, the additional load is $\rho_{\ell}=T_{\ell}^{-1}$, and setting a shorter training period results in a much higher overhead for the system. However, we can note that extremely long periods such as $T_{\ell}=400$ also lead to a slower convergence: in this cumulative plot, convergence is achieved when the reward trend becomes linear.

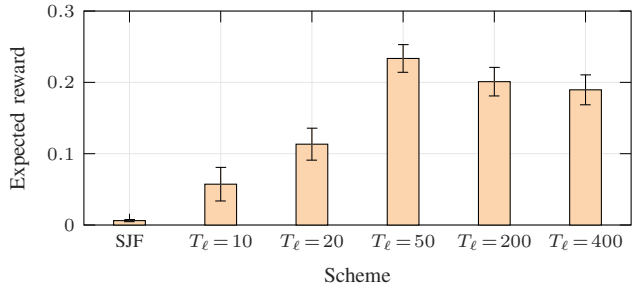
\begin{figure}[t]
\centering
    \begin{tikzpicture}

\pgfplotstableread{
T   avg err
1   0.006162    0.0014785
2   0.057240    0.023542930456
3   0.113333    0.02241450
4   0.233528    0.019415779
5   0.200983    0.020013857
6   0.189537    0.02097185
}\loadedtable;

\begin{axis}[%
width=\fwidth,
height=\ffheight,
ybar,
tick align=inside,
bar width=12pt,
legend style={legend cell align=left, fill opacity=1, draw opacity=1, text opacity=1, legend columns=5, align=left, draw=white!15!black, font=\tiny, at={(0.5, 0.02)}, anchor=south},
xlabel style={font=\footnotesize\color{white!15!black}},
ylabel style={font=\footnotesize\color{white!15!black}},
tick label style={font=\scriptsize\color{white!15!black}},
xmajorgrids,
ymajorgrids,
xmin=0.5,
xmax=6.5,
xtick={1,2,3,4,5,6},
xticklabels={SJF,$T_{\ell}\!=\!10$,$T_{\ell}\!=\!20$,$T_{\ell}\!=\!50$,$T_{\ell}\!=\!200$,$T_{\ell}\!=\!400$},
xlabel={Scheme},
ymin=0,
ymax=0.3,
ylabel={Expected reward},
axis background/.style={fill=white}
]

    \addplot+[style={black,fill={white!50!color4}},error bars/.cd, y dir=both, y explicit] table[x=T, y=avg, y error=err] {\loadedtable};

    \end{axis}
\end{tikzpicture}
\caption{Expected reward after convergence as a function of the training period $T_{\ell}$ in the static scenario.\label{fig:conv_reward_T}}\vspace{-0.3cm}
\end{figure}

We can also explicitly evaluate performance after convergence, considering $N_{\text{test}}=100$ episodes in which the agent exploits the learned policy without further learning.
In an ideal scenario, where the cost of learning has no impact on the system, training as often as possible is highly beneficial.
On the other hand, Fig.~\ref{fig:conv_reward_T} shows that, while increasing the frequency of training is beneficial up to a point, allowing the agent to learn a better policy, setting $T_{\ell}=10$ or $T_{\ell}=20$ leads to a significant decrease in performance.
Initially, less frequent training is more convenient in terms of rewards, as the \gls{mec} scheduler can allocate more resources to users even with unrefined strategies, as the left side of Fig.~\ref{fig:cumulative_T} shows. However, in the long run, less frequent training is inefficient because a suboptimal policy continues to be applied. The optimal working point depends on the specific characteristic of the environment to be optimized and, thus, $T_{\ell}$ must be tuned specifically for each application and scenario.

\begin{figure}[t]
\centering
\begin{tikzpicture}
\begin{axis}[%
width=\fwidth,
height=\fheight,
legend style={legend cell align=left, fill opacity=0.6, draw opacity=1, text opacity=1, legend columns=2, align=left, draw=white!15!black, font=\scriptsize, at={(0.99, 0.02)}, anchor=south east},
xlabel style={font=\footnotesize\color{white!15!black}},
ylabel style={font=\footnotesize\color{white!15!black}},
tick label style={font=\scriptsize\color{white!15!black}},
xmajorgrids,
ymajorgrids,
xmin=0,
xmax=1000,
xlabel={Episode},
ymin=-0.25,
ymax=0.15,
ylabel={Cumulative reward},
axis background/.style={fill=white}
]
\addplot [color=color0, mark=o, mark repeat=50, mark options={solid}]
table[x=Episode,y=SJF]{./fig/comp_cumulative.csv};
\addlegendentry{SJF}

\addplot [color=color1, mark=triangle, mark repeat=50, mark options={solid}]
table[x=Episode,y=pts]{./fig/comp_cumulative.csv};
\addlegendentry{PTS}

\addplot [color=color2, mark=square, mark repeat=50, mark options={solid}]
table[x=Episode,y=ats]{./fig/comp_cumulative.csv};
\addlegendentry{ATS}

\addplot [color=color3, mark=x, mark repeat=50, mark options={solid}]
table[x=Episode,y=ideal]{./fig/comp_cumulative.csv};
\addlegendentry{Ideal}

\end{axis}
\end{tikzpicture}%
\caption{Cumulative reward during the training for all policies in the static scenario.\label{fig:cumulative_all}}\vspace{-0.3cm}
\end{figure}
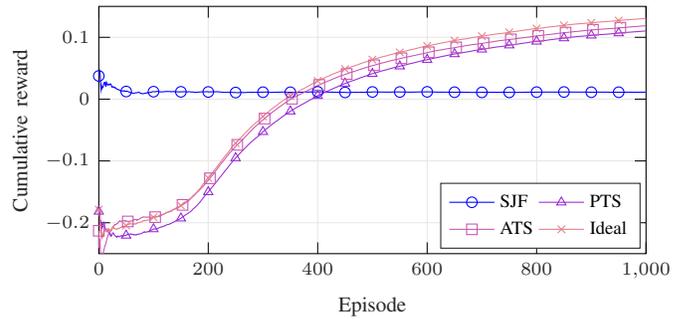

This balance can then be struck by using the \gls{ats} policy, which can almost approach the performance of ideal \gls{drl} method, as Fig.~\ref{fig:cumulative_all} shows.
\gls{ats} also converges faster than \gls{pts}: while all \gls{drl} schemes are initially at a disadvantage with respect to \gls{sjf}, which does not need to insert training jobs in the server, \gls{pts} requires more than $400$ episodes to overcome the performance deficit.
Instead, \gls{ats} outperforms \gls{sjf} approximately $50$ episodes earlier than \gls{pts}, with only a short delay with respect to the ideal training case.
This dual advantage, i.e., faster convergence and higher performance, shows that an intelligent strategy can significantly mitigate the cost of learning problem.

We remark that, unlike our previous work on this field, \gls{ats} does not need application-specific tuning.
This is because the allocation of training jobs affects the state of the system and does not modify the agent's action space, allowing the direct use of Q-value estimates to tune the training process.
In other words, \gls{ats} can figure out the states in which inserting a training job would lead to excessive performance degradation by directly looking at Q-values. This key feature makes \gls{ats} an environment-agnostic method that can be readily generalized to other scenarios affected by the cost of learning.

\begin{figure}[t]
\centering
    \begin{tikzpicture}

\pgfplotstableread{
T   avg err
1    0.006162   0.001478544  
2    0.218928   0.003078585
3    0.243087   0.001435276
4    0.251746   0.001537118
}\loadedtable;

\begin{axis}[%
width=\fwidth,
height=\ffheight,
ybar,
tick align=inside,
bar width=12pt,
legend style={legend cell align=left, fill opacity=1, draw opacity=1, text opacity=1, legend columns=5, align=left, draw=white!15!black, font=\tiny, at={(0.5, 0.02)}, anchor=south},
xlabel style={font=\footnotesize\color{white!15!black}},
ylabel style={font=\footnotesize\color{white!15!black}},
tick label style={font=\scriptsize\color{white!15!black}},
xmajorgrids,
ymajorgrids,
xmin=0.5,
xmax=4.5,
xtick={1,2,3,4},
xticklabels={SJF,PTS,ATS,Ideal},
xlabel={Scheme},
ymin=0,
ymax=0.3,
ylabel={Expected reward},
axis background/.style={fill=white}
]

    \addplot+[style={black,fill={white!50!color4}},error bars/.cd, y dir=both, y explicit] table[x=T, y=avg, y error=err] {\loadedtable};

    \end{axis}
\end{tikzpicture}
\caption{Expected reward after convergence for all the policies in the static scenario.}\label{fig:conv_reward_all}\vspace{-0.3cm}
\end{figure}
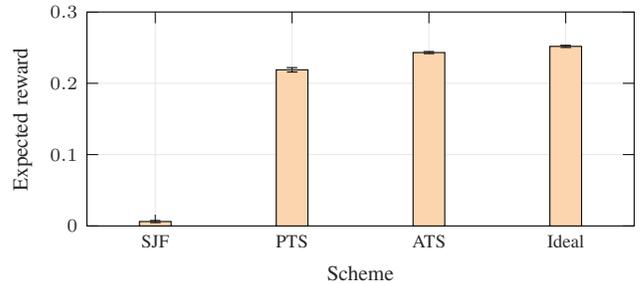

\subsection{Dynamic scenario}
As we discussed in Sec.~\ref{sec:cost}, our framework is particularly suited to non-stationary environment, where the learning agent has to adapt dynamically following a \gls{cl} approach.
We then consider a a dynamic version of the \gls{mec} scenario, where the average load increases linearly from $0.1$ to $0.3$ over the course of $N_{\text{train}}=1500$ episodes. This emulates a scenario in which the distribution of job requests changes with the number or type of users connected to the \gls{bs}.

In this case, the policy obtained by the ideal \gls{drl} approach is pre-trained considering a fixed load of $\rho=0.1$ and then implemented in the dynamic environment without further training.
Due to its lack of adaptation, we refer to this approach as the \textit{fixed strategy}.
In contrast, the policies obtained by the \gls{pts} and \gls{ats} approaches continue to interact with the environment, adapting to variations in the value of $\rho$ and maintaining a fixed exploration rate $\varepsilon=0.1$.

\begin{figure}[t]
\centering
\begin{tikzpicture}
\begin{axis}[%
width=\fwidth,
height=\fheight,
legend style={legend cell align=left, fill opacity=0.6, draw opacity=1, text opacity=1, legend columns=3, align=left, draw=white!15!black, font=\scriptsize, at={(0.99, 0.02)}, anchor=south east},
xlabel style={font=\footnotesize\color{white!15!black}},
ylabel style={font=\footnotesize\color{white!15!black}},
tick label style={font=\scriptsize\color{white!15!black}},
xmajorgrids,
ymajorgrids,
xmin=0,
xmax=1490,
xlabel={Episode},
ymin=-0.05,
ymax=0.2,
ylabel={Reward gap},
axis background/.style={fill=white}
]

\addplot [color=color1, mark=triangle, mark repeat=3, mark options={solid}]
table[x=Episode,y=pts]{./fig/continual_cumulative.csv};
\addlegendentry{PTS}

\addplot [color=color2, mark=square, mark repeat=3, mark options={solid}]
table[x=Episode,y=ats]{./fig/continual_cumulative.csv};
\addlegendentry{ATS}

\addplot [color=color4, mark=x, mark repeat=3, mark options={solid}]
table[x=Episode,y=fixed]{./fig/continual_cumulative.csv};
\addlegendentry{Fixed}

\end{axis}
\end{tikzpicture}%
\caption{Reward gap between SJF and the DRL policies during training in the dynamic scenario.\label{fig:dynamic_gap}}
\end{figure}
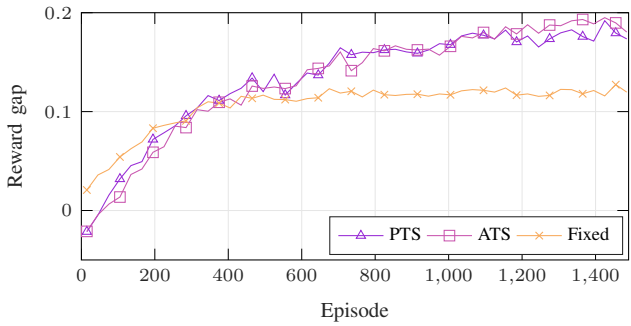

Fig.~\ref{fig:dynamic_gap} shows the relative advantage of the three policies over \gls{sjf}: initially, the fixed strategy outperforms \gls{pts} and \gls{ats}, as they suffer from the overhead due to insertion of training jobs in the \gls{mec} server. However, the gain over \gls{sjf} is relatively small, as setting $\rho=0.1$ makes the system easy to optimize.
The performance gap decreases over time, as the use of a continual learning approach becomes more important, and \gls{ats} and \gls{pts} start outperforming the fixed strategy after $\approx400$ episodes.

\begin{figure}[t]
\centering 
    \begin{tikzpicture}

\pgfplotstableread{
T   avg err
1    0.007252   0.003708267
2    0.228685   0.004128305
3    0.239184   0.004683124
4    0.165362   0.013315165
}\loadedtable;

\begin{axis}[%
width=\fwidth,
height=\ffheight,
ybar,
tick align=inside,
bar width=12pt,
legend style={legend cell align=left, fill opacity=1, draw opacity=1, text opacity=1, legend columns=5, align=left, draw=white!15!black, font=\tiny, at={(0.5, 0.02)}, anchor=south},
xlabel style={font=\footnotesize\color{white!15!black}},
ylabel style={font=\footnotesize\color{white!15!black}},
tick label style={font=\scriptsize\color{white!15!black}},
xmajorgrids,
ymajorgrids,
xmin=0.5,
xmax=4.5,
xtick={1,2,3,4},
xticklabels={SJF,PTS,ATS,Fixed},
xlabel={Scheme},
ymin=0,
ymax=0.3,
ylabel={Expected reward},
axis background/.style={fill=white}
]

    \addplot+[style={black,fill={white!50!color4}},error bars/.cd, y dir=both, y explicit] table[x=T, y=avg, y error=err] {\loadedtable};

    \end{axis}
\end{tikzpicture}
\caption{Expected reward after convergence for all the policies in the dynamic scenario.\label{fig:reward_end}}\vspace{-0.3cm}
\end{figure}
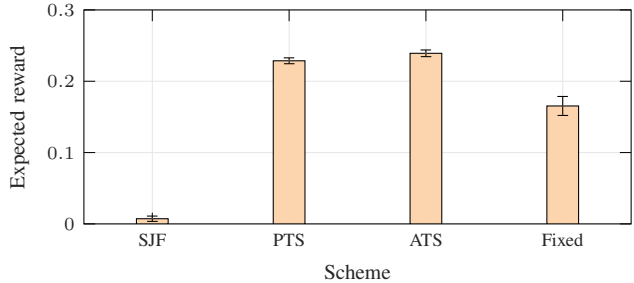

We conducted a final test of $N_{\text{test}}=100$ episodes to evaluate the performance of all the strategies with $\rho=0.3$, after the training phase is over.
Fig.~\ref{fig:reward_end} shows that \gls{ats} reaches a better strategy than \gls{pts}, and both schemes significantly outperform the fixed strategy.
The higher performance of \gls{pts} and \gls{ats} over the benchmarks attests to the need to adopt \gls{cl} approaches, even if the improvement of the agent policy involves a cost for the \gls{mec} server.
These results denote how explicitly considering the cost of learning is fundamental to obtain effective learning strategies in scenarios where training is in direct competition for resources with the users.

\section{Conclusion and Future Work}
\label{sec:conc}
In this work, we proposed a framework to handle the cost of learning in \gls{drl}-based resource allocation problems.
We considered a \gls{mec} system, and designed a framework to mitigate the significant but often overlooked computing overhead associated with learning-based optimization strategies.
Unlike traditional \gls{drl} solutions, our framework considers that the training of learning agents has a direct impact on the system that these agents aim to optimize.
Hence, we proposed an adaptive strategy that identifies the best moments to carry out training, exploiting information derived from the learning process itself.
Our strategy outperforms traditional \gls{drl} approaches and can be readily generalized to other scenarios where the training impact is a critical factor, such as edge network optimization.

Future extension of this work will involve the investigation of the relation between training and exploration decisions, since training effectiveness is unavoidably dependent on the efficiency of the environment exploration. 
Besides, we plan to validate our framework and the proposed training strategy in more scenarios, considering data from real applications. 

\bibliographystyle{IEEEtran}
\bibliography{biblio.bib}

\end{document}